\documentclass[5p]{elsarticle}

\usepackage{amssymb}
\usepackage{booktabs}
\usepackage{pifont}
\newcommand{\xmark}{\ding{55}}%
\usepackage{hyperref}
\usepackage{color}

\usepackage{listings}
\journal{SoftwareX}

\begin{document}
\renewcommand{\labelenumii}{\arabic{enumi}.\arabic{enumii}}

\begin{frontmatter}

\title{humancompatible.detect: a Python Toolkit for Detecting Bias in AI Models}


\author[label1]{Germán M. Matilla}
\author[label1]{Jiří Němeček}
\author[label1]{Illia Kryvoviaz}
\author[label1]{Jakub Mareček}
\address[label1]{Faculty of Electrical Engineering, Czech Technical University in Prague, Prague, \{martige1, nemecj38, kryvoill, jakub.marecek\}@fel.cvut.cz}

\begin{abstract}
\textit{There is a strong recent emphasis on trustworthy AI. In particular, international regulations, such as the AI Act, demand that AI practitioners measure data quality on the input and estimate bias on the output of high-risk AI systems. However, there are many challenges involved, including scalability (MMD) and computability (Wasserstein-1) issues of traditional methods for estimating distances on measure spaces.}

\textit{Here, we present \texttt{humancompatible.detect}, a toolkit for bias detection that addresses these challenges. It incorporates two newly developed methods to detect and evaluate bias: maximum subgroup discrepancy (MSD) and subsampled $\ell_\infty$ distances. It has an easy-to-use API documented with multiple examples. \texttt{humancompatible.detect} is licensed under the Apache License, Version 2.0.}


\end{abstract}


\end{frontmatter}


\label{}

\begin{table*}[!h]
\centering
\begin{tabular}{|l|p{8cm}|p{8cm}|}
\hline
\textbf{Nr.} & \textbf{Code metadata description} & \textbf{Metadata} \\
\hline
C1 & Current code version & 0.1.4 \\
\hline
C2 & Permanent link to code/repository used for this code version & \url{https://github.com/humancompatible/detect/tree/6b3fb21a155d012cb244c9b5b7ead1f8a978038f} \\
\hline
C3  & Permanent link to Reproducible Capsule & none \\
\hline
C4 & Legal Code License   & Apache-2.0 \\
\hline
C5 & Code versioning system used & git \\
\hline
C6 & Software code languages, tools, and services used & Python, pandas, Pyomo, HiGHS, SciPy \\
\hline
C7 & Compilation requirements, operating environments \& dependencies & python $\geq$ 3.10, pandas $\geq$ 2.2 pyomo $\geq$ 6.9, highspy $\geq$ 1.10, scipy $\geq$ 1.16 \\
\hline
C8 & If available Link to developer documentation/manual & \url{https://humancompatible-detect.readthedocs.io/en/latest/} \\
\hline
C9 & Support email for questions & martige1@fel.cvut.cz \\
\hline
\end{tabular}
\caption{Code metadata (mandatory)}
\label{codeMetadata}
\end{table*}



\section{Motivation and significance}

With the increase in practical applications using machine-learning models and AI systems, new ethical problems have appeared. One particular problem is the proliferation (or even strengthening) of societal bias. Infamous examples of AI bias include hiring algorithms \citep{dastin_insight_2018} or recidivism predictors \citep{julia_angwin_machine_2016}.

In response, more and more regulatory frameworks \citep[e.g.,][]{ai-act} and related standards \citep[e.g.,][]{10851955,schwartz_towards_2022} require testing for bias, without any clear agreement on the precise definition of said bias. Indeed, there are many bias measures \citep{mehrabi_survey_2021}.

The lack of a single definition of bias is understandable, considering the wide range of applications of AI systems.
Non-exhaustively:

\begin{itemize}
\item Fairness in (large) language and/or vision models, the currently dominant AI models \cite[e.g.,][]{ahia_magnet_2024,castleman_adultification_2025}.
\item Fairness in rankings: the main related case being job recruitment. See, for instance, \cite{9458629,DBLP:conf/icde/KliachkinPMF24,fageot2025generalizing}.
\item Fairness in recommender systems: such as the ones widely used currently by, e.g., YouTube or Spotify \citep{9458629,liu_when_2025}.
\item Spatial aspects of fairness: In contrast to continuous attributes, such as age and income, the case of discrimination based on location demands special attention  \citep{sacharidis2023auditingspatialfairness,Kyriakopoulos2025}.
\item Fairness for forecasting: Consider the well-known COMPAS dataset. It is traditionally treated as a classification task. A forecasting-based approach, however, obtains the best results currently reported on the dataset \citep{articleQuan}.
\item Fairness in sharing-economy business models: such as the model of companies such as Uber and Cabify \citep{Zhou_2024}.
\end{itemize}

Even when agreeing on a particular definition of bias, several difficulties remain:

\paragraph{Sample Complexity and Computability}
To estimate data quality, i.e., the preexisting bias in the data per se, one generally calculates a distance between the distribution of values in the training data and the general population. There are many such distances on measure spaces: total variation, MMD \citep{gretton2012kernel}, and Wasserstein \citep{vaserstein1969markov}, to name the most popular. However, the sample complexity of these distances varies; that is, the number of samples required to achieve an accurate approximation of the selected distance is very large. Moreover, some of them (esp. Wasserstein-1) are believed to be inapproximable (cf. \cite{lee2023computability,lee2024computability}).

\begin{table*}[t]
    \centering
    \begin{tabular}{rccccc}
        \toprule
        Framework & Detection & Intersections & Low sample complexity & Optimal & Ref \\
        \midrule
        Themis-ml & \xmark & \xmark & - & - & \citep{bantilan2017themismlfairnessawaremachinelearning} \\ 
        FAT Forensics & \checkmark & \xmark & - & - & \cite{sokol2022fat-forensics} \\ 
        Fairlearn & \xmark & \checkmark & - & - & \citep{Weerts_Fairlearn_Assessing_and_2023} \\
        AI Fairness 360 & \checkmark & \checkmark & ? & \xmark & \citep{aif360-oct-2018} \\ 
        FairTest & \checkmark & \checkmark & \checkmark & \xmark & \citep{tramer2015fairtest} \\ 
        \midrule
        \texttt{humancompatible.detect} & \checkmark & \checkmark & \checkmark & \checkmark & here \\
        \bottomrule
    \end{tabular}
    \caption{Comparison to common Python fairness frameworks in four dimensions. Whether they allow for detecting bias (not just measuring bias of a predefined group), if they allow for considering group intersections natively, whether some of those detection methods have low sample complexity, and whether the methods have (optimality) guarantees. When no detection method is available, we omit the last two dimensions. The most similar, FairTest, focuses on identifying protected subgroups using a tree structure that aligns with the predictor's output. Importantly, none of the other frameworks consider the \texttt{MSD} distance or the subsampled $\ell_\infty$ norm.}
    \label{tab:frameworks}
\end{table*}

\paragraph{Choice of Protected Attributes and Intersectional Aspects}
The sample complexity is also clearly tied to the number of protected subgroups. In the case of studying the data quality, one may need to test the distance between the distribution of values in the training data and the general population for each protected subgroup.
The number of protected subgroups may thus enter linearly in the overall sample complexity, while being exponential in the number of protected attributes.

For example, consider
The California Consumer Privacy Act, which lists a number of possible protected attributes and their values to be protected.
The number of protected subgroups can go up to 5,772,124,799 on US Census data \citep{nemecek_bias_2025}.
Estimating the bias for each such protected subgroup is clearly challenging.

\paragraph{Other fairness toolkits}
There is a variety of existing toolkits, of which only some allow for bias detection. Overall are mostly concerned with either training fair models or fairness evaluation. We provide an overview of various toolkits in Table \ref{tab:frameworks}. There are additional toolkits focused on visualiation (e.g., What-If Tool), however, we ommit those from the comparison since they are visual tools rather than evaluation toolkits.

Themis-ml \cite{bantilan2017themismlfairnessawaremachinelearning} and Fairlearn \cite{Weerts_Fairlearn_Assessing_and_2023} do not allow for detecting (most) biased subgroups, and detection part of FAT Forensics \cite{sokol2022fat-forensics} does not allow for the study of intersections of protected groups. AI Fairness 360 \cite{aif360-oct-2018} contains only the MDSS method for the detection of intersectional bias, which uses a non-standard definition of bias and it's sample complexity is unclear. It is also a probabilistic method. FairTest \cite{tramer2015fairtest} contains methods with low sample complexity, however these only test how aligned are protected attributes to the prediction, evaluating correlations rather than standard bias measures. They form a greedy tree, separating groups into intersectional subgroups, without guarantees on optimality.  

\section{Software description}

\begin{figure*}[t]
    \centering
    \includegraphics[width=0.7\linewidth]{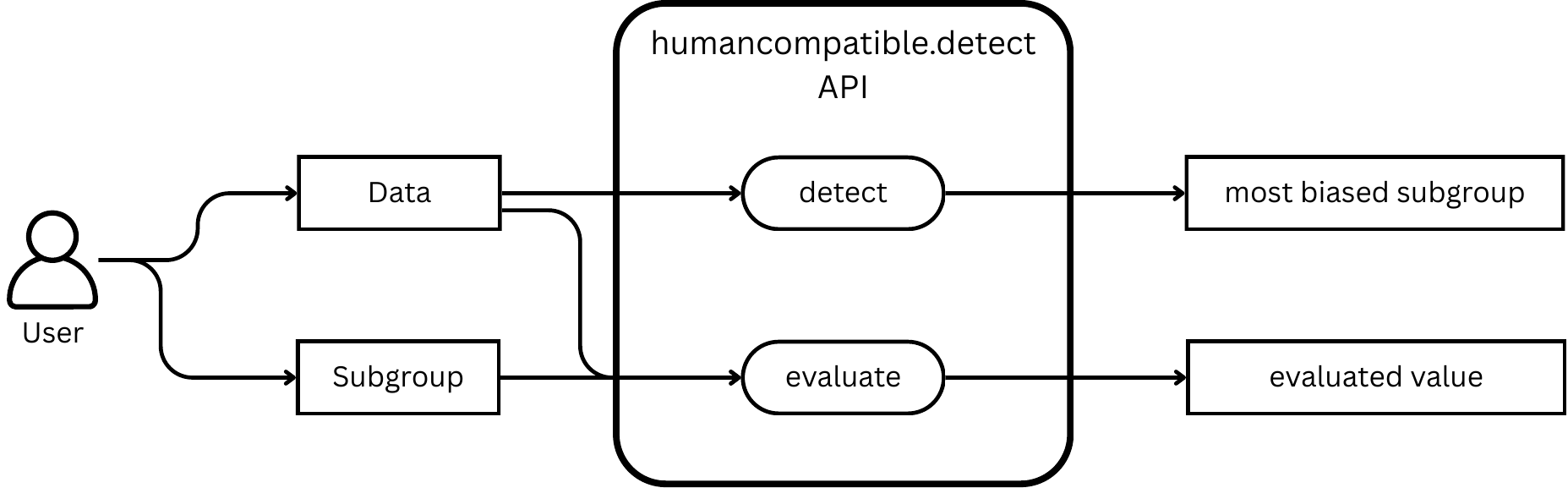}
    \caption{A diagram of possible uses of the \texttt{humancompatible.detect} toolkit. In addition to evaluating bias of a protected group, it allows for detection of the most biased subgroup given the dataset.}
    \label{fig:user_case_diag}
\end{figure*}

We introduce \texttt{humancompatible.detect}
, an open-source toolkit of methods for bias detection. Designed for ease of use, its simple API and guided examples are accessible to non-expert users. It is available on PyPI\footnote{https://pypi.org/project/humancompatible-detect/}. The toolkit allows for reliable detection of intersectional bias and evaluation of group bias even when the number of samples is low. Fig \ref{fig:user_case_diag} illustrates the use of the toolkit.

\subsection{Software architecture}

As per Figure \ref{fig:user_case_diag}, the two main toolkit objectives are \texttt{detect} and \texttt{evaluate}. The first one spots the most biased subgroup in the data, while the second informs us about the bias against a given subgroup within the dataset. The data can be passed in multiple ways.

\paragraph{Input data} 
There are three modes of input:

\begin{itemize}
\item In-memory tables (pandas DataFrame). The dataset is passed in as the input data \texttt{X} and the binary target column \texttt{y}.
\item CSV file, with a designated target column. Requiring the \texttt{path} to this data and the name of the \texttt{column}.
\item Two-sample comparison, where the goal is to identify subgroups whose representations differs most between two distributions, represented by the datasets. Both datasets must share the same columns. One simply passes two pandas DataFrames \texttt{X1} and \texttt{X2} as arguments.
\end{itemize}

\paragraph{Detect bias}
\begin{figure*}[]
    \centering
    \includegraphics[width=\linewidth]{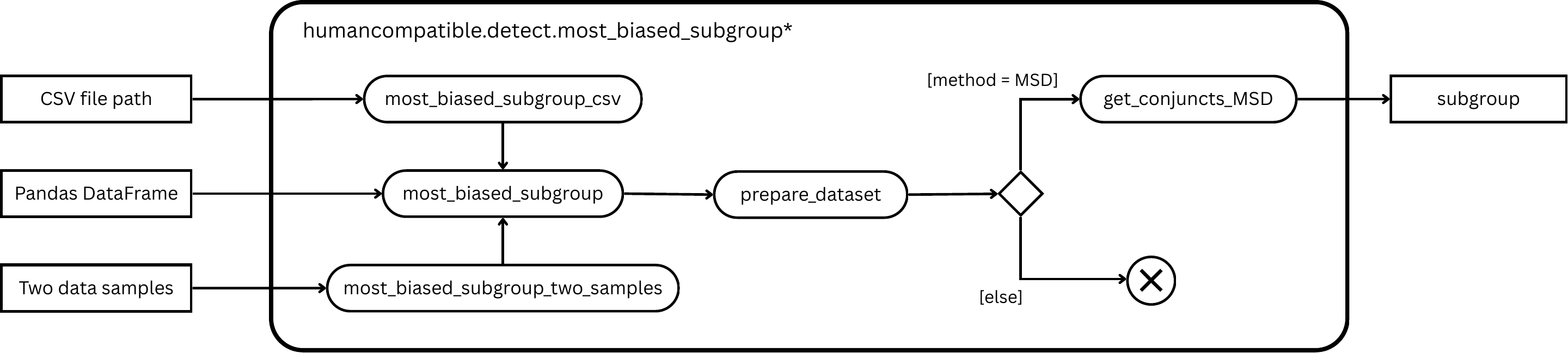}
    \caption{\texttt{detect\_bias} activity diagram. Depending on the input mode used, different functions can be run, always calling \texttt{most\_biased\_subgroup}. After that, the data preprocessing takes places with \texttt{prepare\_dataset}. Subsequently, the program stops if a method different from MSD has been selected, otherwise outputs the most affected subgroup as a result. This allows for easy extensibility. More details on the \texttt{prepare\_dataset} are in Figure \ref{fig:prepare_diag}.}
    \label{fig:detect_diag}
\end{figure*}

To detect bias in data the toolkit alows to use MSD. It can check the exponentially growing number of dataset subgroups and spot the most biased subgroup in the dataset. It returns a simple conjunctive rule representation of a subgroup (e.g.,“AGE between 18-25 AND SEX = Male”). The toolkit can be easily extended to allow for other methods, such as SPSF \cite{kearns_preventing_2018}.
See Figure \ref{fig:detect_diag} for a graphical description.

\paragraph{Evaluate bias}
\begin{figure*}[t]
    \centering
    \includegraphics[width=\linewidth]{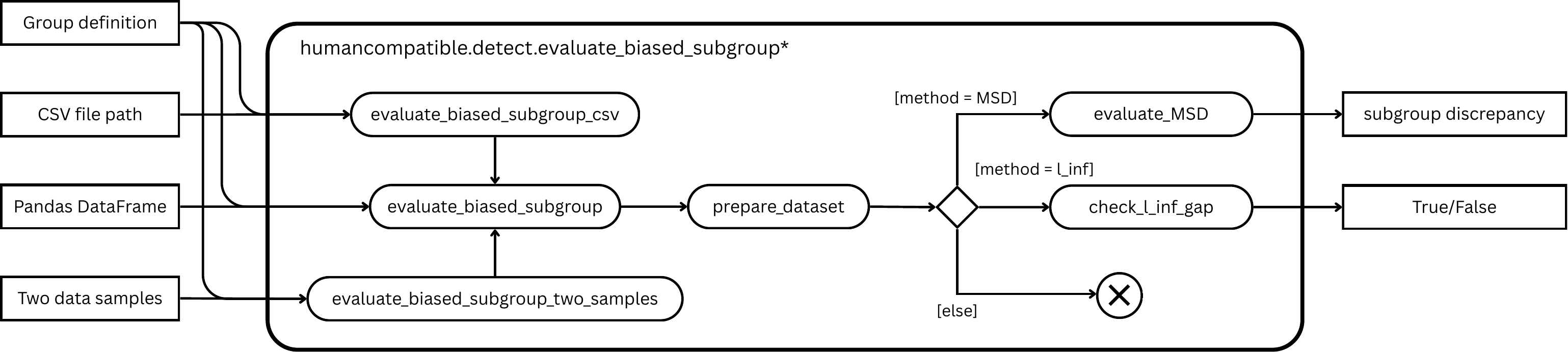}
    \caption{\texttt{evaluate\_bias} activity diagram. Both input and preprocessing are identical to \texttt{detect\_bias}, with the addition of keyword arguments specifying the (sub)group (Group definition) and potentially other arguments (e.g., the $\Delta$ threshold). The output can either be a value (e.g., subgroup discrepancy) or a boolean value whether bias is above a given threshold (e.g., in subsampled $\ell_\infty$). More details on the \texttt{prepare\_dataset} are in Figure \ref{fig:prepare_diag}.}
    \label{fig:eval_diag}
\end{figure*}

Independent of learning which subgroup is the most (dis)advantaged within the data, we can quantify the bias for any subgroup. The MSD method measures the subgroup discrepancy of a given subgroup, which is specified as a conjunction of attributes. Additionally, with our second method, subsampled $\ell_\infty$, we can test whether the bias in a particular subgroup is greater or smaller than a value of our choice. For this task, both methods require additional parameters to be passed as keyword arguments.
See Figure \ref{fig:eval_diag} for a graphical visualization of the process.

\begin{figure*}[t]
    \centering
    \includegraphics[width=0.8\linewidth]{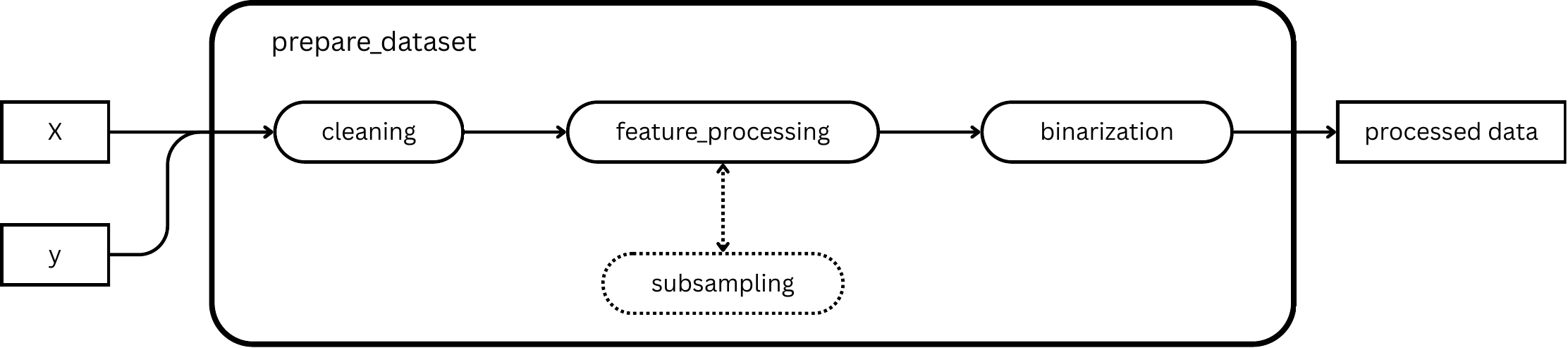}
    \caption{\texttt{prepare\_dataset} activity diagram. This function preprocesses the input data in the form of a pandas DataFrame, \texttt{y}, being the target column and \texttt{X}, the rest of the data. It prunes incomplete rows, allows for binarization, and also for specialized feature processing, such as mapping categorical variables into variables with fewer distinct values or custom binning of continuous variables. It also allows for subsampling the data, if passed data is too large. This can speed up the processing without reducing accuracy too much, since implemented methods have low sample complexity.}
    \label{fig:prepare_diag}
\end{figure*}

Internally, both tasks use a routine to preprocess the input data, \texttt{prepare\_dataset}, which is also illustrated in further detail in Fig \ref{fig:prepare_diag}.

For convenience, we provide a unified wrapper for users, \texttt{detect\_and\_score(\ldots)} combining the detection and evaluation into one step.
It takes the data (in either of the three formats), the protected attributes and target specification, and returns the corresponding subgroup description together with the computed bias score (or pass/fail decision, depending on the method).

\subsection{Software functionalities}

We propose a toolkit that implements two approaches to bias estimation. With these, users can (i) discover the most impacted subgroup using Maximum Subgroup Discrepancy, and (ii) check a concrete subgroup against a tolerance using the subsampled $\ell_\infty$ test.

\paragraph{Maximum Subgroup Discrepancy}

When detecting bias, it is crucial to look at all intersections of protected groups, because marginal fairness might be achieved despite exhibiting severe bias when looking at the intersections.
As a type of curse of dimensionality, the number of group intersections (subgroups) grows exponentially with the number of protected attributes. This complicates the evaluation of intersectional bias in multiple ways. One has to face the exponential time complexity of testing each subgroup, while having exponentially fewer and fewer samples. A useful approach to this would be to consider the bias as a distance between two distributions, but as mentioned, most distributional distances have exponential sample complexity.

Maximum Subgroup Discrepancy (MSD) \citep{nemecek_bias_2025} addresses the issues of high sample complexity of intersectional fairness. It is a distributional distance, defined as
$$\mathrm{MSD}(\mu, \nu) = \sup_{S \in \mathcal{S}} |\mu(S) - \nu(S)|,$$
where $\mu$ and $\nu$ are two distributions, $\mathcal{S}$ is the set of all subgroups, and $\mu(S)$ is the probability of sampling a point from the subgroup $S$ given the distribution $\mu$. In simpler terms, MSD looks for the subgroup with the highest difference in probability between two distributions.
The MSD algorithm has linear sample complexity in the number of protected attributes, returns naturally defined subgroups (as conjunctions of feature value pairs), and the group that is guaranteed to have the highest discrepancy for practical dataset sizes with many protected attributes. Using mixed-integer optimization, it obtains the provably optimal solution (i.e. the maximally discrepant subgroup).

\paragraph{Subsampled Distances on Measure Spaces}

Our second approach addresses the same issue regarding protected group intersections in a different manner. In particular, we subsample a test whether the distance between the two distributions is less than a given value ($\Delta$). For a given subgroup, we can check whether the given subgroup faces a bias within the threshold $\Delta$.
In large datasets, we may have to subsample this test, but \cite{matilla2025samplecomplexitybiasdetection} have developed guaranties on the size of the error with a certain probability. Such guaranties are sometimes known as
probably approximately correct (PAC) learning.

\section{Illustrative example}

We illustrate the toolkit on a synthetic example dataset.

There are three columns: \emph{Race} (categorical: {Green, Blue, Purple}), \emph{Age} (categorical buckets: {0-18, 18-30, 30-45, 45-60}), and a binary \emph{Target} (e.g., approval). The dataset is built so that marginal fairness holds, but one intersectional subgroup is disadvantaged, as shown in Figure \ref{fig:intersect_motiv}.

We treat \emph{Race} and \emph{Age} as protected attributes, and use MSD to find the subgroup.

\lstdefinestyle{hcpython}{
  language=Python,
  basicstyle=\ttfamily\small,
  columns=fullflexible,
  showstringspaces=false,
  frame=single,
}
\lstset{style=hcpython}


\begin{lstlisting}[language={Python}]
from humancompatible.detect import detect_and_score

rule, msd_val = detect_and_score(
    csv_path = "01_data.csv",
    target_col = "Target",
    protected_list = ["Race", "Age"],
    method = "MSD",
)
\end{lstlisting}

The toolkit returns a rule as follows (after interpreting the rule):

\begin{lstlisting}[language={Python}]
print("The group is: " +
      " AND ".join(str(r) for _, r in rule))
print("MSD value is:", msd_val)
\end{lstlisting}
\noindent \texttt{The group is: Race = Blue AND Age = 0-18}\\
\noindent \texttt{MSD value is: 0.111111}

After evaluating the bias for the obtained subgroup, we got the MSD value of $0.111\ldots$ (e.g., an 11.1\% gap between the positive-outcome and the negative outcome for the given subgroup).
MSD confirms that \emph{young Blue} individuals are the most disadvantaged group, even though the data looks fair on the marginals. This is why checking intersections matters.

To test the subsampled $\ell_\infty$ norm, we add a new protected attribute \emph{Gender} (categorical: {M, F}) the previous dataset, while keeping the original demographic parity properties.
We consider the concrete subgroup \emph{Gender = F}, and test whether its distribution of positive outcomes differs from the full population by at most a tolerance $\Delta$ in the $\ell_\infty$ norm.

For a threshold $\Delta = 0.125$, the method reports that the most impacted subgroup bias is \emph{greater} than 0.125 (fail). In other words, the maximum difference between the subgroup and the overall positive outcome histograms is over the threshold.

\section{Impact}

\begin{figure}[t]
    \centering
    \includegraphics[width=\linewidth]{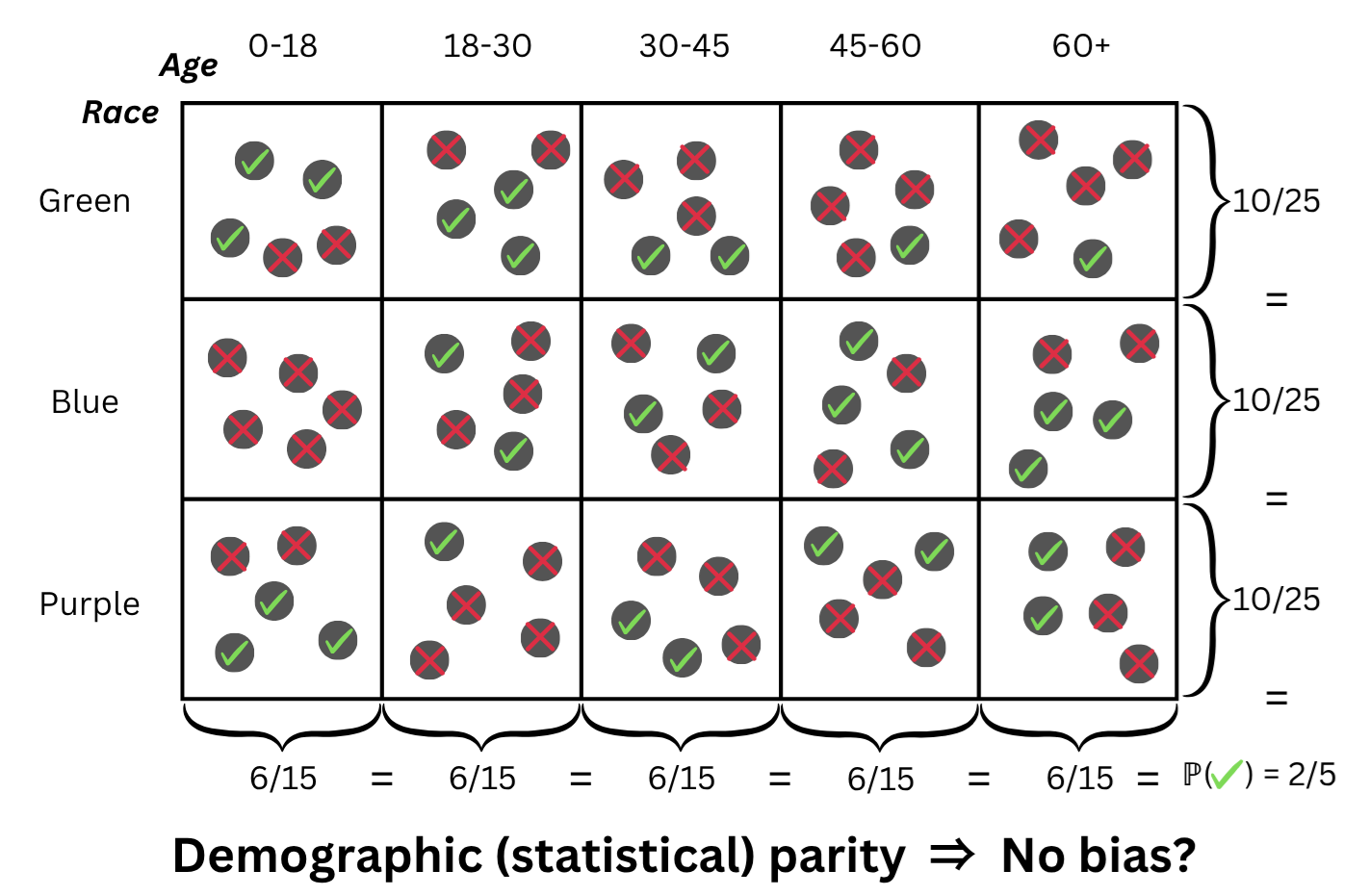}
    \caption{An illustration of why evaluating marginal bias does not suffice. A classifier accepts the same proportion of each age and race group, but rejects the entire subgroup (group intersection) of young blue people. This bias would go undetected unless group intersections are considered. The graphic is a modification of Figure 1 in \cite{nemecek_bias_2025}.}
    \label{fig:intersect_motiv}
\end{figure}

Even in the best-studied case of fairness of classification, most of the proposed bias measures, such as statistical parity or equality of opportunity, focus on protected groups in isolation. This approach often misses the intersections of groups or requires one to sift through an exponential number of group intersections (subgroups). Considering all subgroups is crucial, as illustrated by Figure \ref{fig:intersect_motiv}, where data is marginally fair, but intersectionally biased.

Another view of bias evaluation in cases when many subgroups need to be considered is bias detection. That is, detecting where bias exists in the data, rather than measuring it entirely over all protected features. Detecting such a biased subgroup allows focusing on the particular subgroup \cite[e.g.,][]{sharma_facilitating_2024}. This way, one can address the most significant biases individually through mitigation techniques tailored to the particular group, instead of treating all biases the same.

We showed that \texttt{humancompatible.detect} addresses both questions (intersectional fairness and bias detection) by allowing its users to go from tabular data to a concrete, human-readable finding in a few lines. Moreover, its two implemented methods, have an advantage over previously considered approaches such as those based on MMD \citep{gretton2012kernel} and Wasserstein \citep{vaserstein1969markov} distances, in terms of the number of samples required. For MSD, the number of samples is linear in the number of protected attributes, and the sample complexity can be controlled by controlling the error (or uncertainty) in the estimate.

\section{Conclusions}

We have presented a toolkit for bias detection. Crucially, it can consider protected group intersections and provide bias evaluations with guarantees even for smaller datasets. It has an intuitive API and is designed for easy use by practitioners without requiring domain expertise.

\section*{CRediT author statement}
\textbf{Germán M. Matilla:} Software, Visualization, Validation, Writing - Original Draft, \textbf{Jiří Němeček:} Methodology, Software, Writing - Original Draft, Visualization, Writing - Review \& Editing, \textbf{Illia Kryvoviaz:} Software, Validation, Visualization, Writing - Original Draft, \textbf{Jakub Mareček:} Conceptualization, Supervision, Writing - Original Draft, Writing - Review \& Editing.




\bibliographystyle{elsarticle-num}
\bibliography{sample.bib}

@inproceedings{DBLP:conf/icde/KliachkinPMF24,
  author       = {Andrii Kliachkin and
                  Eleni Psaroudaki and
                  Jakub Marecek and
                  Dimitris Fotakis},
  title        = {Fairness in Ranking: Robustness through Randomization without the
                  Protected Attribute},
  booktitle    = {40th International Conference on Data Engineering, {ICDE} 2024 - Workshops,
                  Utrecht, Netherlands, May 13-16, 2024},
  pages        = {201--208},
  publisher    = {{IEEE}},
  year         = {2024},
  url          = {https://doi.org/10.1109/ICDEW61823.2024.00032},
  doi          = {10.1109/ICDEW61823.2024.00032},
  bibsource    = {dblp computer science bibliography, https://dblp.org}
}

@article{lee2023computability,
  title={Computability of optimizers},
  author={Lee, Yunseok and Boche, Holger and Kutyniok, Gitta},
  journal={IEEE Transactions on Information Theory},
  volume={70},
  number={4},
  pages={2967--2983},
  year={2023},
  publisher={IEEE}
}

@INPROCEEDINGS{9458629,

  author={Pitoura, Evaggelia and Stefanidis, Kostas and Koutrika, Georgia},

  booktitle={2021 IEEE 37th International Conference on Data Engineering (ICDE)}, 

  title={Fairness in Rankings and Recommenders: Models, Methods and Research Directions}, 

  year={2021},

  volume={},

  number={},

  pages={2358-2361},

  doi={10.1109/ICDE51399.2021.00265}}

@misc{sacharidis2023auditingspatialfairness,
      title={Auditing for Spatial Fairness}, 
      author={Dimitris Sacharidis and Giorgos Giannopoulos and George Papastefanatos and Kostas Stefanidis},
      year={2023},
      eprint={2302.12333},
      archivePrefix={arXiv},
      primaryClass={cs.LG},
      url={https://arxiv.org/abs/2302.12333}, 
}

@INPROCEEDINGS{fageot2025generalizing,
  title={Generalizing while preserving monotonicity in comparison-based preference learning models},
  author={Fageot, Julien and Blanchard, Peva and Bareilles, Gilles and Hoang, L{\^e}-Nguy{\^e}n},
  note={arXiv preprint arXiv:2506.08616},
booktitle={Thirty-ninth Conference on Neural Information Processing Systems},
  year={2025}
}

@INPROCEEDINGS{Kyriakopoulos2025,
      title={PROMIS: A Post-Processing Framework for Mitigating Spatial Bias}, 
      author={Dimitris Kyriakopoulos and Dimitris Sacharidis and Giorgos Giannopoulos and Dimitris Gunopulos and Theodore Dalamagas},
      year={2025},
      booktitle = {Proceedings of the 32nd ACM International Conference on Advances in Geographic Information Systems}
}

@article{articleQuan,
author = {Zhou, Quan and Marecek, Jakub and Shorten, Robert},
year = {2023},
month = {04},
pages = {1247-1280},
title = {Fairness in Forecasting of Observations of Linear Dynamical Systems},
volume = {76},
journal = {Journal of Artificial Intelligence Research},
doi = {10.1613/jair.1.14050}
}

@inproceedings{Zhou_2024,
   title={Closed-Loop View of the Regulation of AI: Equal Impact across Repeated Interactions},
   url={http://dx.doi.org/10.1109/ICDEW61823.2024.00029},
   DOI={10.1109/icdew61823.2024.00029},
   booktitle={2024 IEEE 40th International Conference on Data Engineering Workshops (ICDEW)},
   publisher={IEEE},
   author={Zhou, Quan and Ghosh, Ramen and Shorten, Robert and Mareček, Jakub},
   year={2024},
   month=may, pages={176–181} }

@inproceedings{nemecek_bias_2025,
    address = {Toronto ON Canada},
    title = {Bias {Detection} via {Maximum} {Subgroup} {Discrepancy}},
    isbn = {979-8-4007-1454-2},
    url = {https://dl.acm.org/doi/10.1145/3711896.3736857},
    doi = {10.1145/3711896.3736857},
    language = {en},
    urldate = {2025-09-19},
    booktitle = {Proceedings of the 31st {ACM} {SIGKDD} {Conference} on {Knowledge} {Discovery} and {Data} {Mining} {V}.2},
    publisher = {ACM},
    author = {Němeček, Jiří and Kozdoba, Mark and Kryvoviaz, Illia and Pevný, Tomáš and Mareček, Jakub},
    month = aug,
    year = {2025},
    pages = {2174--2185},
}

@article{dastin_insight_2018,
    chapter = {World},
    title = {Insight - {Amazon} scraps secret {AI} recruiting tool that showed bias against women},
    url = {https://www.reuters.com/article/world/idUSKCN1MK0AG/},
    language = {en-US},
    urldate = {2025-04-01},
    journal = {Reuters},
    author = {Dastin, Jeffrey},
    month = oct,
    year = {2018},
}

@article{mehrabi_survey_2021,
    title = {A {Survey} on {Bias} and {Fairness} in {Machine} {Learning}},
    volume = {54},
    issn = {0360-0300},
    url = {https://dl.acm.org/doi/10.1145/3457607},
    doi = {10.1145/3457607},
    number = {6},
    urldate = {2025-04-01},
    journal = {ACM Comput. Surv.},
    author = {Mehrabi, Ninareh and Morstatter, Fred and Saxena, Nripsuta and Lerman, Kristina and Galstyan, Aram},
    month = jul,
    year = {2021},
    pages = {115:1--115:35},
}

@misc{julia_angwin_machine_2016,
    title = {Machine {Bias}},
    url = {https://www.propublica.org/article/machine-bias-risk-assessments-in-criminal-sentencing},
    language = {en},
    urldate = {2023-04-29},
    journal = {ProPublica},
    author = {{Julia Angwin} and {Jeff Larson} and {Lauren Kirchner} and {Surya Mattu}},
    month = may,
    year = {2016},
}

@article{10851955,
    title = {{IEEE} standard for algorithmic bias considerations},
    doi = {10.1109/IEEESTD.2025.10851955},
    journal = {IEEE Std 7003-2024},
    author = {{IEEE Standards Association}},
    year = {2025},
    pages = {1--59},
}

@misc{ai-act,
    title = {Regulation ({EU}) 2024/1689 of the {European} {Parliament} and of the {Council}. of 13 {June} 2024 laying down harmonised rules on artificial intelligence and amending {Regulations} ({EC}) {No} 300/2008, ({EU}) {No} 167/2013, ({EU}) {No} 168/2013, ({EU}) 2018/858, ({EU}) 2018/1139 and ({EU}) 2019/2144 and {Directives} 2014/90/{EU}, ({EU}) 2016/797 and ({EU}) 2020/1828 ({Artificial} {Intelligence} {Act})},
    urldate = {2025-02-10},
    author = {{Regulation (EU) 2024/1689}},
    year = {2024},
    note = {Place: OJ L, 2024/1689, 12.7.2024},
}

@misc{matilla2025samplecomplexitybiasdetection,
      title={Sample Complexity of Bias Detection with Subsampled Point-to-Subspace Distances},
      author={M. Matilla, Germán and Mareček, Jakub},
      year={2025},
      eprint={2502.02623},
      archivePrefix={arXiv},
      primaryClass={cs.LG},
      url={https://arxiv.org/abs/2502.02623v1},
}

@inproceedings{sharma_facilitating_2024,
    address = {New York, NY, USA},
    series = {{CHI} '24},
    title = {Facilitating {Self}-{Guided} {Mental} {Health} {Interventions} {Through} {Human}-{Language} {Model} {Interaction}: {A} {Case} {Study} of {Cognitive} {Restructuring}},
    isbn = {979-8-4007-0330-0},
    shorttitle = {Facilitating {Self}-{Guided} {Mental} {Health} {Interventions} {Through} {Human}-{Language} {Model} {Interaction}},
    url = {https://dl.acm.org/doi/10.1145/3613904.3642761},
    doi = {10.1145/3613904.3642761},
    urldate = {2025-09-24},
    booktitle = {Proceedings of the 2024 {CHI} {Conference} on {Human} {Factors} in {Computing} {Systems}},
    publisher = {Association for Computing Machinery},
    author = {Sharma, Ashish and Rushton, Kevin and Lin, Inna Wanyin and Nguyen, Theresa and Althoff, Tim},
    month = may,
    year = {2024},
    pages = {1--29},
}

@misc{schwartz_towards_2022,
    title = {Towards a {Standard} for {Identifying} and {Managing} {Bias} in {Artificial} {Intelligence}},
    url = {https://tsapps.nist.gov/publication/get_pdf.cfm?pub_id=934464},
    language = {en},
    publisher = {Special Publication (NIST SP), National Institute of Standards and Technology, Gaithersburg, MD},
    author = {Schwartz, Reva and Vassilev, Apostol and Greene, Kristen K. and Perine, Lori and Burt, Andrew and Hall, Patrick},
    month = mar,
    year = {2022},
    doi = {https://doi.org/10.6028/NIST.SP.1270},
}

@incollection{lee2024computability,
  title={Computability of optimizers for AI and data science},
  author={Lee, Yunseok and Boche, Holger and Kutyniok, Gitta},
  booktitle={Handbook of Numerical Analysis},
  volume={25},
  pages={359--388},
  year={2024},
  publisher={Elsevier}
}

@article{gretton2012kernel,
  title={A kernel two-sample test},
  author={Gretton, Arthur and Borgwardt, Karsten M and Rasch, Malte J and Sch{\"o}lkopf, Bernhard and Smola, Alexander},
  journal={The Journal of Machine Learning Research},
  volume={13},
  number={1},
  pages={723--773},
  year={2012},
  publisher={JMLR. org}
}

@article{vaserstein1969markov,
  title={Markov processes over denumerable products of spaces, describing large systems of automata},
  author={Vaserstein, Leonid Nisonovich},
  journal={Problemy Peredachi Informatsii},
  volume={5},
  number={3},
  pages={64--72},
  year={1969},
  publisher={Russian Academy of Sciences, Branch of Informatics, Computer Equipment and~…}
}

@article{tramer2015fairtest,
  title={FairTest: Discovering Unwarranted Associations in Data-Driven Applications},
  author={Tramer, Florian and Atlidakis, Vaggelis and Geambasu, Roxana and Hsu, Daniel
          and Hubaux, Jean-Pierre and Humbert, Mathias and Juels, Ari and Lin, Huang},
  journal={arXiv preprint arXiv:1510.02377},
  year={2015}
}

@misc{aif360-oct-2018,
    title = "{AI Fairness} 360:  An Extensible Toolkit for Detecting, Understanding, and Mitigating Unwanted Algorithmic Bias",
    author = {Rachel K. E. Bellamy and Kuntal Dey and Michael Hind and
	Samuel C. Hoffman and Stephanie Houde and Kalapriya Kannan and
	Pranay Lohia and Jacquelyn Martino and Sameep Mehta and
	Aleksandra Mojsilovic and Seema Nagar and Karthikeyan Natesan Ramamurthy and
	John Richards and Diptikalyan Saha and Prasanna Sattigeri and
	Moninder Singh and Kush R. Varshney and Yunfeng Zhang},
    month = oct,
    year = {2018},
    url = {https://arxiv.org/abs/1810.01943}
}

@article{Weerts_Fairlearn_Assessing_and_2023,
author = {Weerts, Hilde and Dudík, Miroslav and Edgar, Richard and Jalali, Adrin and Lutz, Roman and Madaio, Michael},
journal = {Journal of Machine Learning Research},
title = {{Fairlearn: Assessing and Improving Fairness of AI Systems}},
url = {http://jmlr.org/papers/v24/23-0389.html},
volume = {24},
year = {2023}
}

@misc{bantilan2017themismlfairnessawaremachinelearning,
      title={Themis-ml: A Fairness-aware Machine Learning Interface for End-to-end Discrimination Discovery and Mitigation}, 
      author={Niels Bantilan},
      year={2017},
      eprint={1710.06921},
      archivePrefix={arXiv},
      primaryClass={cs.CY},
      url={https://arxiv.org/abs/1710.06921}, 
}

@article{sokol2022fat-forensics,
  title={{FAT Forensics}: {A} {Python} Toolbox for Algorithmic Fairness,
         Accountability and Transparency},
  author={Sokol, Kacper and Santos-Rodriguez, Raul and Flach, Peter},
  journal={Software Impacts},
  pages={100406},
  year={2022},
  publisher={Elsevier}
}

@inproceedings{liu_when_2025,
    address = {New York, NY, USA},
    series = {{FAccT} '25},
    title = {When {Collaborative} {Filtering} is not {Collaborative}: {Unfairness} of {PCA} for {Recommendations}},
    isbn = {979-8-4007-1482-5},
    shorttitle = {When {Collaborative} {Filtering} is not {Collaborative}},
    url = {https://doi.org/10.1145/3715275.3732190},
    doi = {10.1145/3715275.3732190},
    urldate = {2026-01-07},
    booktitle = {Proceedings of the 2025 {ACM} {Conference} on {Fairness}, {Accountability}, and {Transparency}},
    publisher = {Association for Computing Machinery},
    author = {Liu, David and Baek, Jackie and Eliassi-Rad, Tina},
    month = jun,
    year = {2025},
    pages = {2974--2990},
}

@inproceedings{castleman_adultification_2025,
    address = {New York, NY, USA},
    series = {{FAccT} '25},
    title = {Adultification {Bias} in {LLMs} and {Text}-to-{Image} {Models}},
    isbn = {979-8-4007-1482-5},
    url = {https://doi.org/10.1145/3715275.3732178},
    doi = {10.1145/3715275.3732178},
    urldate = {2026-01-07},
    booktitle = {Proceedings of the 2025 {ACM} {Conference} on {Fairness}, {Accountability}, and {Transparency}},
    publisher = {Association for Computing Machinery},
    author = {Castleman, Jane and Korolova, Aleksandra},
    month = jun,
    year = {2025},
    pages = {2751--2767},
}

@article{ahia_magnet_2024,
    title = {{MAGNET}: {Improving} the {Multilingual} {Fairness} of {Language} {Models} with {Adaptive} {Gradient}-{Based} {Tokenization}},
    volume = {37},
    shorttitle = {{MAGNET}},
    url = {https://proceedings.neurips.cc/paper_files/paper/2024/hash/5572bc595de865c1450868fd5391e9c5-Abstract-Conference.html},
    doi = {10.52202/079017-1514},
    language = {en},
    urldate = {2026-01-07},
    journal = {Advances in Neural Information Processing Systems},
    author = {Ahia, Orevaoghene and Kumar, Sachin and Gonen, Hila and Hofmann, Valentin and Limisiewicz, Tomasz and Tsvetkov, Yulia and Smith, Noah A.},
    month = dec,
    year = {2024},
    pages = {47790--47814},
}

@inproceedings{kearns_preventing_2018,
    title = {Preventing {Fairness} {Gerrymandering}: {Auditing} and {Learning} for {Subgroup} {Fairness}},
    shorttitle = {Preventing {Fairness} {Gerrymandering}},
    url = {https://proceedings.mlr.press/v80/kearns18a.html},
    language = {en},
    urldate = {2025-04-16},
    booktitle = {Proceedings of the 35th {International} {Conference} on {Machine} {Learning}},
    publisher = {PMLR},
    author = {Kearns, Michael and Neel, Seth and Roth, Aaron and Wu, Zhiwei Steven},
    month = jul,
    year = {2018},
    note = {ISSN: 2640-3498},
    pages = {2564--2572},
}

\end{document}